\documentclass[pdflatex,sn-mathphys-num]{sn-jnl}

\usepackage{graphicx}%
\usepackage{multirow}%
\usepackage{amsmath,amssymb,amsfonts}%
\usepackage{amsthm}%
\usepackage{mathrsfs}%
\usepackage[title]{appendix}%
\usepackage{xcolor}%
\usepackage{textcomp}%
\usepackage{manyfoot}%
\usepackage{booktabs}%
\usepackage{cleveref}%
\usepackage{algorithm}%
\usepackage{algorithmicx}%
\usepackage{algpseudocode}%
\usepackage{listings}%

\theoremstyle{thmstyleone}%
\theoremstyle{thmstyletwo}%

\theoremstyle{thmstylethree}%

\begin{document}

\title[A General Peg-in-Hole Assembly Policy Based on Domain Randomized Reinforcement Learning]{A General Peg-in-Hole Assembly Policy Based on Domain Randomized Reinforcement Learning}


\author[1]{\fnm{Xinyu} \sur{Liu}}\email{xinl@mmmi.sdu.dk}

\author[1]{\fnm{Aljaz} \sur{Kramberger}}

\author[1]{\fnm{Leon} \sur{Bodenhagen}}

\affil[1]{\orgdiv{The Maersk Mc-Kinney Moller Institute}, \orgname{University of Southern Denmark}, \orgaddress{\city{Odense}, \country{Denmark}}}


\abstract{Generalization is important for peg-in-hole assembly, a fundamental industrial operation, to adapt to dynamic industrial scenarios and enhance manufacturing efficiency. While prior work has enhanced generalization ability for pose variations, spatial generalization to six degrees of freedom (6-DOF) is less researched, limiting application in real-world scenarios. This paper addresses this limitation by developing a general policy GenPiH using Proximal Policy Optimization(PPO) and dynamic simulation with domain randomization. The policy learning experiment demonstrates the policy's generalization ability with nearly 100\% success insertion across over eight thousand unique hole poses in parallel environments, and sim-to-real validation on a UR10e robot confirms the policy’s performance through direct trajectory execution without task-specific tuning.}




\maketitle

\section{Introduction}\label{sec1}

With rapid advancements in robotics and artificial intelligence (AI), robots are increasingly deployed in various industrial applications, particularly for automating mass production to enhance production efficiency. Peg-in-hole assembly, which is the fundamental operation of robotic assembly, becomes important and attracts research attention \cite{valavanis1991general,jiang2022review}. Recently, the integration of DRL has enabled notable improvements in policy's generalization ability in this task \cite{park2017compliance,beltran2020variable}. Typically, DRL-based policy is trained to process observations, such as target poses, and output corresponding actions, like joint positions or end-effector movements, to guide the robot in task execution \cite{Elguea2023review}. These policies can adapt to changing working scenarios, making them highly effective in addressing environmental uncertainties. This adaptability is particularly beneficial for flexible and customized manufacturing. Previous studies have successfully applied DRL to peg-in-hole assembly, achieving strong performance in generalizing across varying object poses\cite{lee2020making, SeqPolicy}. However, most research focuses on various planar positions, and the peg is already roughly aligned with the hole. Generalization to various spatial pose with 6 degrees of freedom (DOF) has been less explored.

This study addresses the gap by employing a DRL-based learning method to implement peg-in-hole assembly with variations in the hole poses. A simulation environment including Universal Robot UR10e robot and Cranfield benchmark \cite{Cranfield} models is constructed using NVIDIA's Isaac Sim and Isaac Lab \cite{IsaacSim} as the extension for training the assembly policies. The PPO algorithm is used for policy learning, as it is known for its stability and capability to process continuous data. The trained policy is then deployed in a real-world setup for experimental validation. In this paper, we outline the following contributions:
\begin{itemize}
    \item A general-purpose simulation environment for training assembly policies.
    \item An assembly policy with generalization ability to various spatial hole pose.
\end{itemize}
\section{Related Work}\label{sec2}

\subsection{Deep Reinforcement Learning-Based Robotic Control} 

DRL-based methods have recently become increasingly popular in robotics because of their capacity to process high-dimensional and continuous data. It combines deep learning and reinforcement learning methods, using neural networks as the policy to process high-dimensional observation and output continuous actions. It has been applied for complex real-world robotic manipulation tasks where the observation space stretches over multiple dimensions, and action space requires continuous values\cite{kroemer2021review}. Generalization of robot tasks is a popular research topic, which gained traction in the past with research of statistical methods \cite{kramberger2017}, whereas today, novel simulation-based DRL methods are taking the forefront. Unlike traditional one-off motion planning or control methods designed for single robotics tasks, learned policies—when appropriately trained—can generalize across multiple tasks in unstructured environments and unknown scenarios. Actor-critic (AC) algorithms are one of the most popular DRL algorithm types, including Proximal policy optimization (PPO) \cite{PPO} and Soft Actor-Critic (SAC) \cite{SAC}. PPO is more stable with generalized advantage estimator \cite{GAE} while SAC is more efficient in complex tasks due to entropy regularization. In this paper, the PPO algorithm is used to learn the PiH task.

\subsection{Peg-in-Hole Assembly}

Peg-in-hole assembly is a fundamental industrial operation that has been researched for decades. These studies mainly focus on assembly performance, including assembly precision and generalization ability to various working scenarios \cite{Elguea2023review}. For peg-in-hole and similar tasks, researchers apply the Deep Deterministic Policy Gradient (DDPG) \cite{DDPG} algorithms to train the control policy for precise timber assembly \cite{apolinarska2021robotic}. In terms of the generalization ability of the policy, it focuses on the object geometry and position in the robot workspace. Some work uses multimodal observations to extract the pose features from various objects, improving the generalization ability on object geometry \cite{lee2020making, SeqPolicy}. While others use the DRL algorithms to train assembly policy with generalization ability to various poses \cite{Jin2021contact}, which is also this paper's focus.

\section{Methodology}

\begin{figure}
    \centering
    \includegraphics[width=0.9\textwidth]{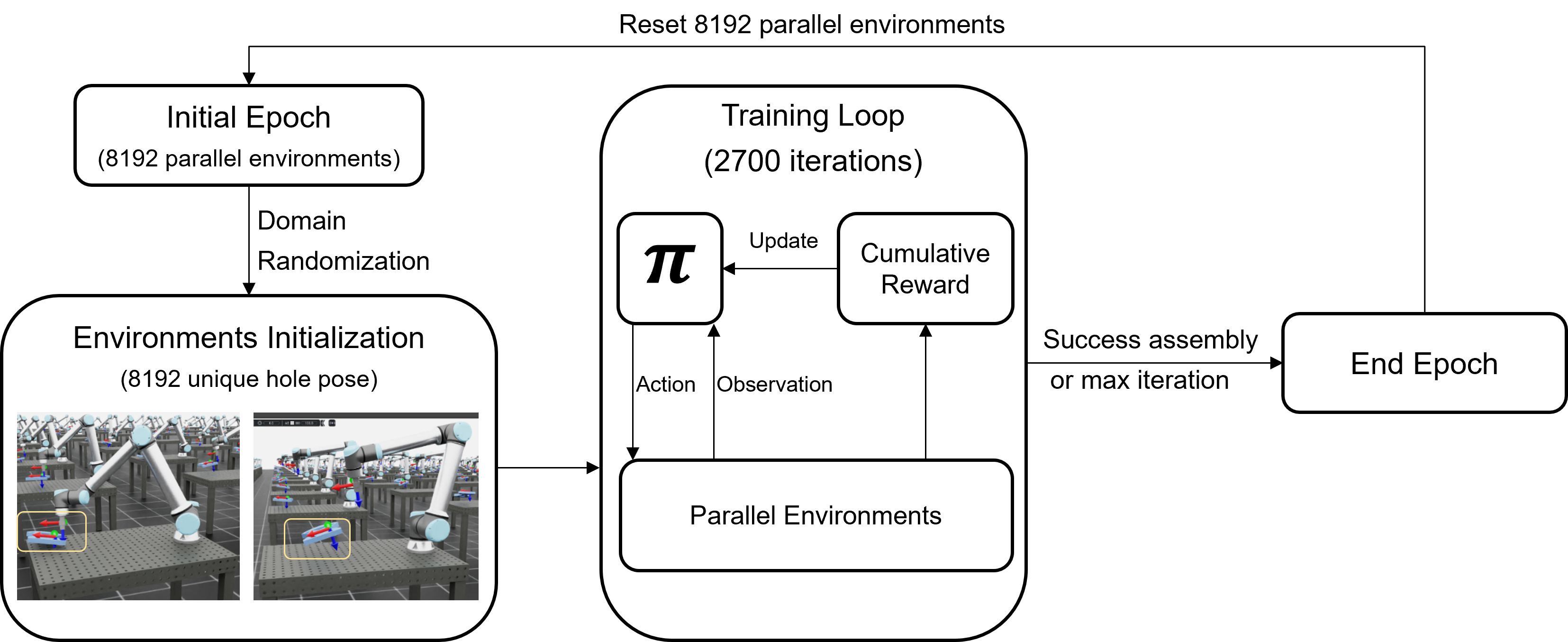}
    \caption{Training pipeline.}
    \label{fig: methodology}
\end{figure}

This work exploits dynamic simulation with domain randomization, and DRL approaches to train the general assembly policy in the UR10e workspace. The training pipeline is shown in \cref{fig: methodology}.

Each training epoch begins by initializing 8,192 parallel environments defined in Isaac Labs, each with a unique hole pose. All environments provide observations to the learning policy, which outputs corresponding actions. After applying the actions, environments update to new states, and rewards are calculated to update the policy. The epoch ends when every environment has successful insertion or reaches the maximum defined loop iterations. Then, all environments reset and enter the next epoch.

\subsection{Dynamic Simulation}
Dynamic simulation provides a framework for developing and testing control strategies in the virtual environment before transitioning to the real robotic setup. In this work, the simulation contains 8,192 parallel environments running simultaneously to generate data for policy learning. Domain randomization in terms of randomizing the hole pose is applied from the range shown in \cref{tab: Hole Pose Range} for every environment to improve the policy's generalization ability.

\begin{table}
\centering
\caption{\centering Hole Pose Range}
\setlength{\tabcolsep}{6.6mm}
\begin{tabular}{cc}
    \hline \rule{0pt}{8pt}
    Variables & Range \\ 
    \hline
    X  & $[-0.2, 0.2]m$ \\
    Y  & $[-0.26, 0.26]m$\\
    Z   & $[0.0, 0.16]m$\\
    RPY   & $[-25, 25][deg]$\\
    \hline
\end{tabular}
\label{tab: Hole Pose Range}%
\end{table}%

The six-dimensional action $\mathbf{a}$ corresponds to the six joint positions $w_i, i=1,...,6$ used to control the robot. The observation space contains the hole (target) pose that is represented with Cartesian coordinate position $\mathbf{p_{hole}}$ and orientation defined as a quaternion $\mathbf{q_{hole}}$, last output action $\mathbf{a}_{t-1}$, where $t$ is the training time-step.

\subsection{Peg-in-hole Assembly Policy Learning}

In this work, the policy is structured as a two-layer neural network, with each layer consisting of 64 neurons. It processes collected observations and determines actions for the next learning step for the simulated robot.

\subsubsection{Proximal Policy Optimization}
The policy is trained with the PPO algorithm, with the training objective formulated as:
\begin{equation}
R = \frac{\pi_\theta(\mathbf{a}|\mathbf{s})}{\pi_{\theta_{\text{old}}}(\mathbf{a}|\mathbf{s})} 
\end{equation}
\begin{equation}
\mathop{\arg\max}\limits_{\theta} {\mathbb{E}} \left[ \min \left( R * A_t^{GAE}, \text{clip}\left(R, 1 - \epsilon, 1 + \epsilon \right) A_t^{GAE} \right)  \right]
\end{equation}
where the objective is to maximize the reward expectation in the entire training epoch. $A_t^{GAE}$ is the Generalized Advantage Estimator \cite{GAE} valued in each training time-step $t$ in the training epoch, $\pi_\theta$ represents the assembly policy, and $\epsilon$ is a predefined parameter which is 0.2 to clip the update ratio $R$, the probability of taking action $\mathbf{a}$ under environment state $\mathbf{s}$ between the new policy and old policy, in a stable range, to avoid overfit and maintaining training stability.

The reward function combines dense and sparse reward together for efficient training. In the dense reward function, the distance between the peg and target pose is calculated in each step as shown below:

\begin{equation}
    d_{q} = \Vert \log(\mathbf{q}_{hole}*\overline{\mathbf{q}}_{peg}) \Vert
    \label{orientation distance}
\end{equation}
\begin{equation}
    d_{p} = \Vert \mathbf{p}_{hole} - \mathbf{p}_{peg} \Vert
    \label{position distance}
\end{equation}
where $\mathbf{q}_{hole}$ and $\mathbf{q}_{peg}$ represent the orientation of the peg and hole defined as unit quaternion $\mathbf{Q}=[v+\mathbf{u}]$, $\mathbf{q}_{hole}, \mathbf{q}_{peg} \in S^3$, where $S^3$ is a unit sphere in $\mathbb{R}^4$, furthermore, $\mathbf{p}_{peg}$ and $\mathbf{p}_{hole}$ are their Cartesian coordinates. The position distance $d_p$ is calculated in the Cartesian space while the difference between two unit quaternions $d_q$ is calculated with the $log$ function; more information on quaternion math can be found in \cite{Ude2014}.

Then, the dense reward is calculated based on the orientation and position distance:
\begin{equation}
    d_{hp} = \sqrt{{d_{q}}^2 + {d_{p}}^2}
    \label{distance function}
\end{equation}
\begin{equation}
    r_{dense} = 1 - \tanh(d_{hp})
    \label{dense reward}
\end{equation}
where the $d_{hp}$ is the pose distance. As the peg approaches and aligns with the hole $d_{hp}$ decreases, the reward $r_{dense}$ increases.

The sparse rewards are provided when the alignment and insertion conditions are met:
\begin{equation}
     r_{parse} = \begin{cases} 2.6 & \text{if $d_{q} < 0.05rad$}\\ 
     10 & \text{if $d_{q} < 0.05rad$ \& $d_{p} < 0.003m$ }  \\
     0 & \text{otherwise}\end{cases}
\end{equation}
where parse reward $r_{parse}$ is provided when $d_q$ and $d_p$ fall below the predefined threshold, indicating a successful alignment or insertion of the peg. Please note the peg pose is defined in the hole frame.

\section{Experiment}
The learned policy was evaluated in simulation and replayed on the real robot to assess the quality of the generated trajectories. Seven comparison experiments were conducted to evaluate policy performance for each DOF individually and for all six DOFs combined.

\subsection{Policy Learning}
\begin{figure}
    \centering
    \includegraphics[width=1\textwidth]{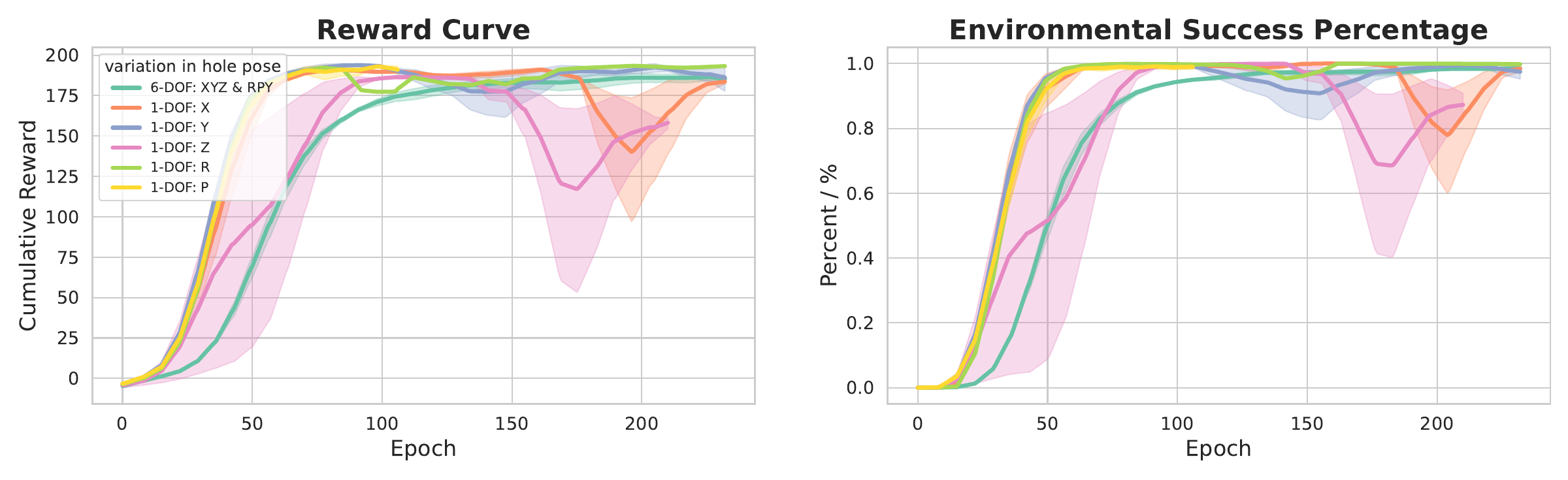}
    \caption{Policy learning metrics.}
    \label{fig: reward curve}
\end{figure}

The policy learning result with two metrics is shown in \cref{fig: reward curve}. The reward curve represents the cumulative reward per epoch, providing an overview of the policy's convergence trend and stability. Meanwhile, the environmental success percentage shows the task-specific performance, defined as:

\begin{equation}
\text{environmental success percentage} = \frac{N_{\text{success}}}{N_{\text{total}}} \times 100
\end{equation}
where $N_{\text{total}}$ is fixed at 8,192, accounting for all of the training environments, and the percentage is calculated at the end of each epoch with $N_{\text{success}}$, representing the proportion of environments achieving successful insertions.

The results indicate efficient policy learning, with quick convergence and nearly 100\% success across environments within 100 epochs. While the performance remained stable across most experiments, performance drops are observed in the X and Z DOFs position experiments after convergence. These drops are primarily due to the algorithm's exploration mechanism, where low-probability actions are occasionally taken to explore potential better solutions. While this behavior temporarily reduces performance, it aids in discovering optimal policies. The final result remains unaffected, as the best-performing model-rather than the last one-is selected for subsequent experiments. Additionally, the training length varies between experiments, usually a few extra epochs after convergence.
 
\subsection{Sim-to-Real Experiment}

The real experiment is conducted with a UR10e robot and custom-designed objects with the same measurements as the Cranfield benchmark. To verify the policy's performance, the hole pose in the real setup is given to the policy to generate the trajectory in simulation, and then the trajectory is replayed in the real setup.

\begin{figure}
    \centering
    \includegraphics[width=0.88\textwidth]{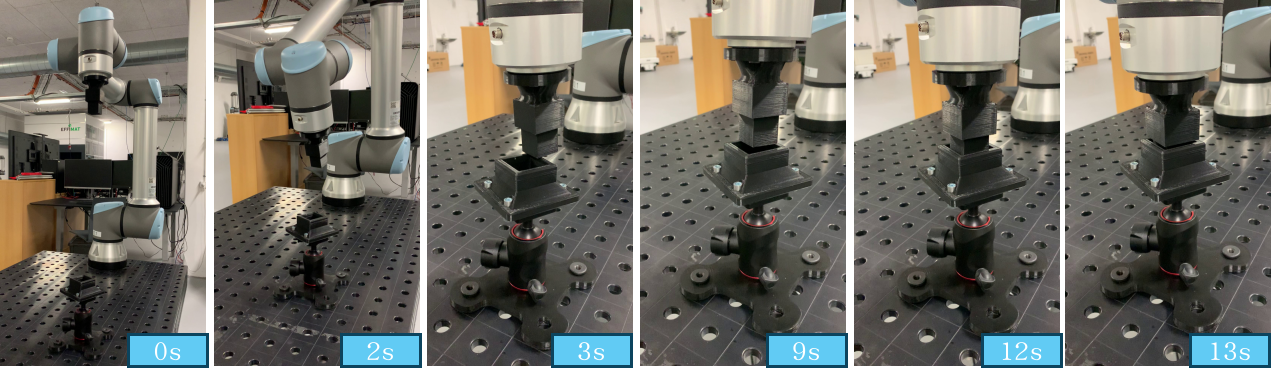} \\
    \hspace{0.1in}
    \includegraphics[width=0.88\textwidth]{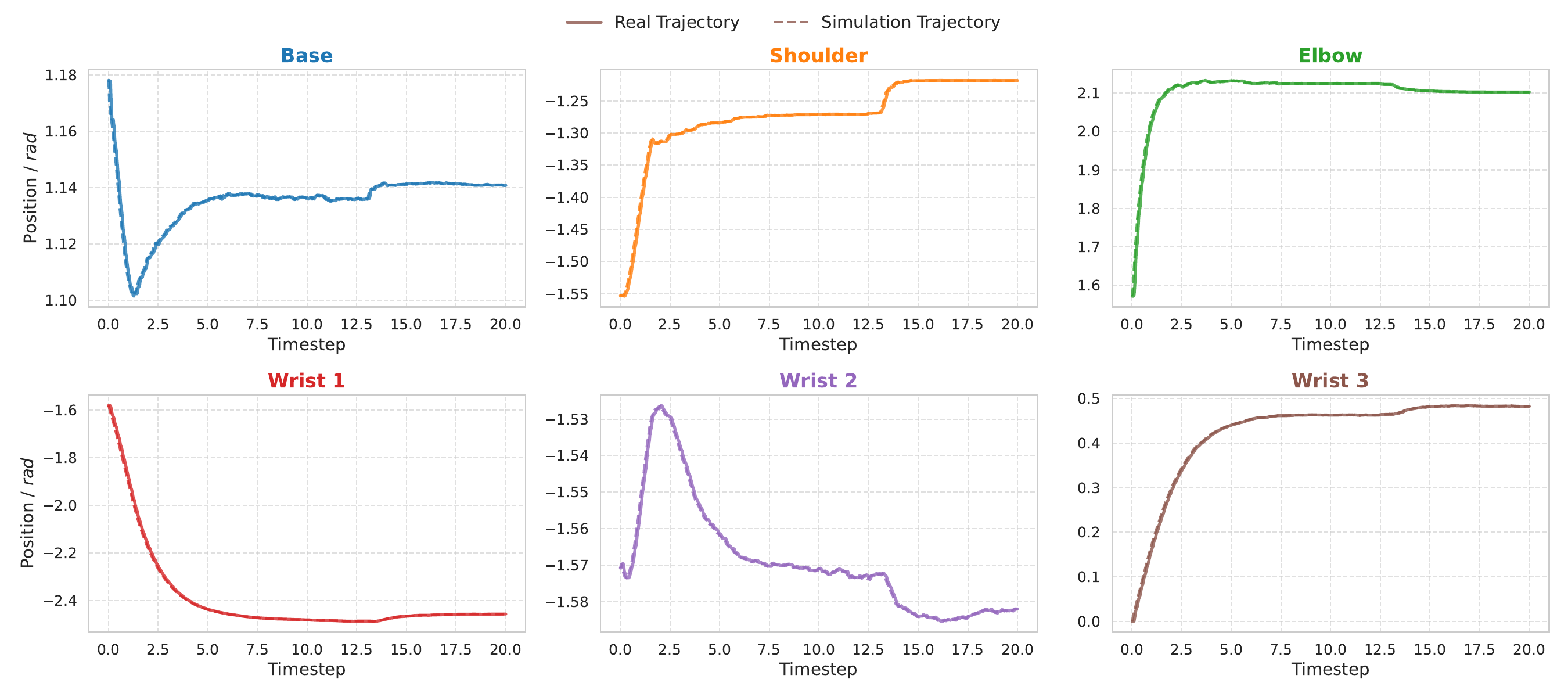}
    \caption{The assembly process in real experiment and joints trajectory.}
    \label{fig: sim2real assmbly process}
\end{figure}

The hole is fixed on the table randomly within the robot workspace. With the digital angle meter, the orientation in the robot base frame could be measured quickly. In the real setup, the table frame and robot base frame are not aligned, therefore, the peg is manually positioned into the hole to measure the accurate relative position directly in the robot base frame. Afterward, the pose is used in the simulation, where the policy generates a trajectory to insert the peg into the hole. This trajectory is executed and verified on the real setup. The assembly process and corresponding real joints trajectory is shown in \cref{fig: sim2real assmbly process}.

The initial joint positions are [67.5, -90, 90, -90, -90, 0]degree, and the target TCP position and Euler angles in robot base frame are [-0.131, -0.703, 0.198]m and [0, 0, 25]degree. The entire assembly process takes 13 seconds. At first, the robot moves toward the hole within 3 seconds and then slows down for precise alignment before inserting the peg. The sim-to-real experiment validates the assembly policy performance, although there are slight vibrations on the robot, especially on the base and wrist joints.

\section{Conclusion}

This study provides a policy that can generalize to various spatial hole pose with 6-DOF for the peg-in-hole assembly task. The policy learning process is efficient, achieving rapid convergence to optimal performance within 100 epochs while maintaining stability. In the sim-to-real experiment, the trajectory generated by the policy exhibits rapid alignment and precise peg insertion, validating the policy's effectiveness.

However, the overall trajectory is not optimal, as it includes redundant motions, such as unnecessary base joint swings and floating peg movements during insertion. Future work will focus on optimizing the joint trajectory by introducing additional constraints or new configurations to generate smoother, more efficient trajectories.

\section*{Acknowledgment}

This work has been funded by the EU project Fluently (Grant agreement ID: 101058680) and supported by the Industry 4.0 lab at the University of Southern Denmark.


\end{document}